\pdfoutput=1

\documentclass[11pt]{article}

\usepackage{EMNLP2022}

\usepackage{times}
\usepackage{latexsym}
\usepackage{amssymb}
\usepackage[smallerops]{newtxmath}
\usepackage{stmaryrd}
\usepackage[T1]{fontenc}

\usepackage[utf8]{inputenc}

\usepackage{microtype}
\usepackage{booktabs}
\usepackage{multirow}
\usepackage{enumitem}
\usepackage{makecell}
\usepackage{adjustbox}
\usepackage{textcomp}
\usepackage{subcaption}
\DeclareMathAlphabet{\mathcal}{OMS}{cmsy}{m}{n}
\DeclareMathAlphabet{\mathbb}{U}{msb}{m}{n}
\usepackage[scaled=0.85]{FiraMono}

\newcommand*{\courier}{\fontfamily{pcr}\selectfont}

\newcommand{\posr}[1]{\textcolor{ForestGreen}{#1}}
\newcommand{\negr}[1]{\textcolor{red}{#1}}
\title{An Empirical Study on Finding Spans}

\author{Weiwei Gu\textsuperscript{\rm 1\textasteriskcentered{}} \quad Boyuan Zheng\textsuperscript{\rm 2\textasteriskcentered{}} \\ \bf Yunmo Chen\textsuperscript{\rm 2} \quad Tongfei Chen\textsuperscript{\rm 3} \quad Benjamin Van Durme\textsuperscript{\rm 2} \\
\textsuperscript{1}~University of Rochester \quad \textsuperscript{2}~Johns Hopkins University \\
\textsuperscript{3}~Microsoft Semantic Machines \\
\courier{\small wgu7@ur.rochester.edu, \{bzheng12,yunmo,vandurme\}@jhu.edu, tongfei@pm.me}
}

\begin{document}
\renewcommand{\thefootnote}{\fnsymbol{footnote}}
\maketitle
\footnotetext[1]{~Equal contribution.}
\renewcommand{\thefootnote}{\arabic{footnote}}
\begin{abstract}

  We present an empirical study on methods for \emph{span finding}, the selection of consecutive tokens in text for some downstream tasks. We focus on approaches that can be employed in training end-to-end information extraction systems, and find there is no definitive solution without considering task properties, and provide our observations to help with future design choices: 1) a tagging approach often yields higher precision while span enumeration and boundary prediction provide higher recall; 2) span type information can benefit a boundary prediction approach; 3) additional contextualization does not help span finding in most cases.
 
\end{abstract}

\section{Introduction}
Various information extraction (IE) tasks require a \emph{span finding} component, which either directly yields the output or serves as an essential component of downstream linking.  In named entity recognition (NER), spans in text are detected and typed; coreference resolution (RE) requires mention spans; mention spans are linked when performing relation extraction (RE), and in event extraction this also requires detection of \emph{trigger spans}.  In extractive  question answering (QA), a span in a passage is detected to be presented as the answer to a given question. 

\begin{figure}[t]
    \centering
    \includegraphics[width=0.5\textwidth]{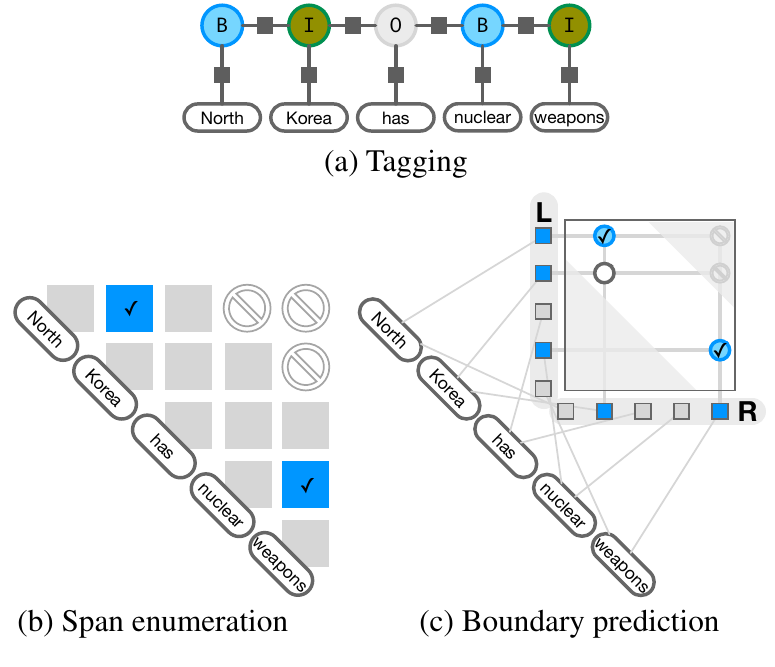}
    \caption{Three common methods for span finding, detecting the mention spans \underline{\emph{North Korea}} and \underline{\emph{nuclear weapons}} in the sentence. ``$\varobslash$'' denotes a candidate span that is too long to be considered.}
    \label{fig:span-finding}
\end{figure}

\newcommand{\Question}{\textbf{Q:~}}
\newcommand{\Answer}{\textbf{A:~}}

\begin{table*}[t]
  \centering\adjustbox{max width=\textwidth}{
  \begin{tabular}{cccc}
    \toprule
     \bf Task & \bf Tagging & \bf Span Enumeration & \bf Boundary Prediction \\
    \midrule
     NER &
       \makecell{ 
         Features + decision trees \small\cite{sekine-etal-1998-decision} \\ 
         Features + CRF \small\cite{mccallum-li-2003-early} \\
         BiLSTM + CRF \small\cite{lample-etal-2016-neural}
       } 
       & \makecell{
         Instance-based \small\cite{ouchi-etal-2020-instance} \\
         \textsc{SpanNer} \small\cite{fu-etal-2021-spanner}
       } 
       & \makecell{
          As QA \small\cite{li-etal-2020-unified}
       } \\ 
     \midrule
     EE  & 
       \makecell{
         JointEE \small Features+CRF \cite{yang-mitchell-2016-joint} \\
         As Cloze \small BERT+CRF \cite{chen-etal-2020-reading}
       } 
       & \makecell{
         \textsc{DyGIE} \small BiLSTM+GNN \cite{luan-etal-2019-general} \\
         \textsc{DyGIE++} \small BERT+GNN \cite{wadden-etal-2019-entity}
      }
       & As QA \small\cite{du-cardie-2020-event}\\
    \midrule
     RE  & 
       \makecell{
         BiLSTM + CRF \small\cite{DBLP:journals/corr/abs-1804-07847}
       }
       & 
       \makecell{
         \textsc{PURE} \small \cite{zhong-chen-2021-frustratingly}\\
       } &
       \makecell{
         \textsc{CasRel} \small\cite{wei-etal-2020-novel}
       } \\
    \midrule
    CR  & 
       Features + MEMM \small\cite{florian-etal-2004-statistical} & 
       Span Ranking \small BiLSTM \cite{lee-etal-2017-end} &
       s2e Coref \small \cite{kirstain-etal-2021-coreference}
    \\
    \midrule
     QA  & 
       \makecell{
         Features + CRF \small\cite{yao-etal-2013-answer} \\ 
         BiLSTM + CRF \small\cite{du-cardie-2018-harvesting}
       } & 
       \makecell{
         \textsc{RaSoR} \small BiLSTM \cite{DBLP:journals/corr/LeeKP016}
       } & 
       \makecell{
         BiDAF \small\cite{SeoKFH17} \\ 
         BERT \small on SQuAD  \cite{devlin-etal-2019-bert}
       }  \\  
    \bottomrule
  \end{tabular}}
  \caption{An overview of various span detecting methods for various tasks in recent NLP literature.}
  \label{tab:overview}
\end{table*}

Following the proliferation of large pre-trained models \cite[\emph{i.a.}]{peters-etal-2018-deep,devlin-etal-2019-bert,RaffelSRLNMZLL20}, recent approaches 
to span finding 
can be roughly divided into three different types: as \emph{tagging}, \emph{span enumeration}, or \emph{boundary prediction} (see \S\ref{sec:background}, \autoref{tab:overview}). 
In this paper, we present an empirical study of these methods and their influence on downstream tasks, hoping to shed light on how to build future NLP systems. 
Specifically, we discuss the design choice of span finding methods and examine two common tricks for improving performance. We answer the following questions:
\begin{enumerate}[leftmargin=*]
    \item \Question What is the best span finding method? Does this hold for various NLP tasks or different pre-trained encoders? \par
    \Answer The choice depends on the downstream task: tagging generally has higher precision, but span enumeration or boundary prediction has higher recall. For most cases, boundary prediction is preferable to span enumeration. Tagging performs much better on a masked language model pretrained encoder (e.g., RoBERTa) than an encoder-decoder pretrained model (e.g., T5).
    \item \Question Does inclusion of mention type information help (e.g., just \texttt{B-PERSON} or \texttt{B} in tagging)? \par
    \Answer For downstream IE tasks, tagging and span enumeration approaches prefer untyped tags, but boundary prediction heavily relies on type information to obtain good performance.
    \item \Question Is an additional contextualization with RNN layers on top of Transformers helpful? Some prior work put an LSTM layer on top of embeddings produced by Transformers \cite[i.a.]{strakova-etal-2019-neural,shibuya-hovy-2020-nested,wang-etal-2021-automated}. Are these necessary? \par
    \Answer Not for RoBERTa (a pretrained encoder-only), but we observe a slight benefit from BiLSTM layers atop T5 (an encoder-decoder).
\end{enumerate}


\section{Background} \label{sec:background}

We assume a pre-trained base model is in place (e.g. BERT \cite{devlin-etal-2019-bert}, T5 \cite{RaffelSRLNMZLL20}). The encoding for each token $x_i$ is denoted as $\mathbf{x}_i \in \mathbb{R}^d$. For further analyses, we focus on RoBERTa \cite{roberta} and T5 since these represent two classes of pretrained models: one with an encoder trained with reconstruction loss; the other with both an encoder and a decoder. 

\paragraph{Tagging}
Span selection can be reduced to a sequence tagging problem, usually under the BIO scheme. Such tagging problems can be modeled using a linear-chain \emph{conditional random field} \cite[CRF; ][]{LaffertyMP01}, with tags either typed or untyped (for example, in NER tags may be labeled with entity types \texttt{B-PERSON}, \texttt{I-LOCATION}, or without,  \texttt{B}, \texttt{I}). In recent work, features input to CRF may be hand-crafted features or predominantly outputs of a neural network. 
Note that reducing span finding to a tagging problem does not allow the system to produce \emph{overlapping} spans. There exist methods to address these (e.g., for the task of \emph{nested} NER), but we leave the analysis of these methods for future work.

\paragraph{Span Enumeration}
\citet{lee-etal-2017-end} first proposed to enumerate all spans (up to length $k$) and predict whether they are entity mentions for coreference resolution. A span embedding $\mathbf{s}_{ij}$ is derived for each span $x[i:j]$, usually a concatenation of the left and right boundary tokens $\mathbf{x}_i, \mathbf{x}_j$, a pooled version of all the tokens between $x_i$ and $x_j$ (usually an attention-weighted sum with a learned global query vector $\mathbf{q}$ \cite{lee-etal-2017-end,lin-ji-2019-attentive}), and optionally some additional manual features $\boldsymbol{\phi}$:
\begin{align}
    a_{k \in \{i, \cdots, j\}} &\propto \exp (\mathbf{q} \cdot \mathbf{x}_k) \nonumber \\
    \mathbf{s}_{ij} &= \left[ \mathbf{x}_i~;~ \mathbf{x}_j~;~  {\displaystyle\sum}_{k=i}^j a_k \mathbf{x}_k ~;~ \boldsymbol{\phi} \right] \nonumber
\end{align}
Such span embedding can be used for both span detection (untyped) and span typing: for detection, a span is given a score indicating whether it is a span of interest by applying a feedforward network $F$ on top of the span embedding:
 \begin{equation}
  \{ x[i : j] \mid F(\mathbf{s}_{ij}) > 0, 0 \le j - i \le k \} \nonumber
 \end{equation}
For typing, one can create a classifier with the span embedding as input, and the set of types plus an $\varepsilon$ type (not a selected span) as the output label set.

\begin{table*}[t]
\resizebox{\textwidth}{!}{%
    \centering
    \begin{tabular}{c|c|ccccccc|c|cc}
      \toprule
         \bf Encoder & \bf Method & \multicolumn{2}{c}{\bf Entities} & \multicolumn{4}{c}{\bf EE} & \bf RE & \bf CR & \multicolumn{2}{c}{\bf QA} \\
         \cmidrule(lr){3-4}\cmidrule(lr){5-8}\cmidrule(lr){9-9}\cmidrule(lr){10-10}\cmidrule(lr){11-12}
        & & Ent-I & Ent-C & Trig-I & Trig-C & Arg-I & Arg-C & F1 & Avg. F1 & EM & F1 \\
      \midrule
        \multirow{3}{*}{RoBERTa\textsubscript{base}}& Tagging & 94.6 & \bf 90.0 & \bf 73.3 &  68.9 & \bf 53.4 & \bf 50.8 &\bf 59.5  &54.5 & 74.4 & 80.8\\
        & Span Enumeration & 94.8 & 89.8 & 72.7 &\bf 69.0 & 22.6 & 16.7 & 13.1 &\bf 72.5 & 68.8 & 73.8 \\
        & Boundary Prediction & \bf 94.9 & 89.9 & 72.3 & 67.7 & 41.6 & 37.4 & 39.6 &71.7 & \bf 77.6 & \bf 85.0\\
      \midrule
        \multirow{3}{*}{$\rm T5_{base}^{enc}$}& Tagging & 94.5 & 88.6 & 60.2 & 39.0 & 24.2 & 22.4 &18.6 & 55.1& 58.1 & 67.3 \\
        & Span Enumeration & 92.0 & 86.2 & 59.2 & 47.8 & 20.2 & 14.1 & 10.2 & 70.4& 62.3 & 66.1\\
        & Boundary Prediction & \bf 95.5 & \bf 90.5 & \bf 70.7 & \bf 68.3 & \bf 39.4 & \bf 35.5 & \bf 37.3 & \bf 70.5 &\bf 70.3 &\bf 76.1\\
      \bottomrule
    \end{tabular}
    }
    \caption{Basic experimental results on downstream tasks that involve mention detection. We report the results of entity extraction and event extraction from ACE05-E\textsuperscript{+} dataset (F-score, \%), relation extraction from ACE05-R dataset (F-score, \%), coreference resolution from OntoNotes dataset, and QA task from SQuAD 2.0 dataset.}
    \label{tab:major_exps}
\end{table*}

\paragraph{Boundary Prediction}
BiDAF \cite{SeoKFH17} introduced a method to select one span from a sequence of text that we term \emph{boundary prediction}. This has been widely adopted following the proliferation of work based on pretrained models such as BERT for QA. 

In boundary prediction, two vectors $\mathbf{l}, \mathbf{r}$ over tokens are computed to indicate whether a token is a left or right boundary of a span:
\begin{equation}
l_i \propto \exp (\mathbf{q}_{\rm L} \cdot \mathbf{x}_i);\quad r_j \propto \exp (\mathbf{q}_{\rm R} \cdot \mathbf{x}_j) \nonumber
\end{equation}
To determine the most likely span, one selects $[\arg\max_i l_i, \arg\max_j r_j]$. 

This method can be extended to the case where the model does not select any span \cite{devlin-etal-2019-bert}. A special [\textsc{cls}] token may be prepended to the sequence of tokens, taking index 0, and the model is trained to select the placeholder span $[0, 0]$ if no span should be selected.

 Boundary prediction can also be extended to select more than one span. In \textsc{CasRel} \cite{wei-etal-2020-novel}, instead of selecting the most likely left and right indices, they select multiple left and right index if their score surpasses a threshold:
 \begin{equation}
         L = \{i \mid l_i \ge \theta \}; \quad R = \{j \mid r_i \ge \theta \} \nonumber
 \end{equation}
 Then a heuristic is used to match these candidate left/right boundaries to select spans. \citet{li-etal-2020-unified} extended the idea: instead of a heuristic, a model $F$ (can be a 2-layer feedforward network, as is used in our experiments) is used to score all candidate spans selected by the threshold (up to length $k$):
 \begin{equation}
  \{ x[i : j] \mid F(\mathbf{s}_{ij}) > 0, i \in L, j \in R, 0 \le j - i \le k \} \nonumber
 \end{equation}
 Since it is the most flexible and heuristic-free, we will focus on the last method in \citet{li-etal-2020-unified}. To modify this span detector into a typed classifier, one can apply the same trick in span enumeration.

\section{Experimental Setup}
We perform all our experiments on RoBERTa\textsubscript{base} and the encoder part of T5\textsubscript{base} (${\rm T5}_{\rm base}^{\rm enc}$). Each number reported is an average of 3 runs with different random seeds. 
Each run is trained with a single Quadro RTX 6000 GPU with 24GB memory.

\paragraph{NER, EE, and RE}
We use the ACE 2005 dataset \citep{ace05} to evaluate model performance on NER, EE, and RE tasks. We follow OneIE \citep{lin-etal-2020-joint} to compile dataset splits and establish the baseline using their released codebase.

For comparable setups across tasks, we disable all the global features which involve complicated cross-subtask interactions and cross-instance interactions that are hard to adapt to other span finding methods. We also disable the additional biaffine entity classifier and event type classifier and use the typing from the span finding module directly in inference time for the experiments in \autoref{tab:major_exps}.
 
For NER, in addition to the standard entity classification F1 (Ent-C), we also report entity identification F1 (Ent-I) to measure how models detect spans. For EE, we use the standard \{trigger (Trig) / argument (Arg)\}-\{identification (I) / classification (C)\} F1 scores. For RE, the standard F1 is used.




\paragraph{Coreference Resolution}
We use the higher-order coreference resolution model \citep{lee-etal-2018-higher} as implemented in AllenNLP \cite{gardner-etal-2018-allennlp} as the baseline for coreference resolution.

We report the average F1 (Avg. F1) of the three common metrics, namely MUC, B$^3$, and CEAF$_{\phi_4}$ on the OntoNotes dataset \cite{ontonotes}.


\paragraph{Extractive QA}
We use the Transformer QA model in BERT as the baseline. 
We evaluate on the dev set of SQuAD 2.0 \citep{rajpurkar-etal-2018-know}, a large-scale reading comprehension dataset containing both answerable and unanswerable questions. We keep the first span in the input sequence for the questions with multiple answer spans and discard the others. 
 We use exact match (EM) and token overlap F1 for QA.

\begin{table*}[t]
\begin{minipage}[t]{0.3\linewidth}
    \centering
    \resizebox{1\linewidth}{!}{%
    \begin{tabular}[t]{lcc}
      \toprule
      \bf Approach & \bf P & \bf R \\
      \midrule
         Tagging & \bf 81.7 & 72.8 \\
         Span Enumeration & 25.5 & \bf 96.1 \\
         Boundary Prediction & 25.5 & \bf 96.0 \\
      \bottomrule
    \end{tabular}}
    \captionsetup{width=.9\linewidth}
    \caption{Breakdown of mention score of each span finding method on OntoNotes dataset. We present the results from the models using RoBERTa\textsubscript{base} encoder.}
    \label{tab:mention_breakdown}
\end{minipage}
\hspace{0.2cm}
\begin{minipage}[t]{.68\linewidth}
\centering
\resizebox{1\textwidth}{!}{%
    \begin{tabular}[t]{l|ccccccc}
      \toprule
        \bf Method & \multicolumn{2}{c}{\bf Entities} & \multicolumn{4}{c}{\bf EE} & \bf RE \\
        \cmidrule(lr){2-3}\cmidrule(lr){4-7}\cmidrule(lr){8-8}
         & Ent-I & Ent-C & Trig-I & Trig-C & Arg-I & Arg-C & F1\\
      \midrule
         Tagging & 94.6 & 89.6 & 72.7 & 69.1  & 54.5 & 52.1 &56.5\\
         \quad - w/o Typing & \posr{+0.4} & \posr{+0.5} & \posr{+0.4} & \negr{\textminus1.1} & \posr{+3.2} & \posr{+3.0} & \posr{+9.6}\\
      \midrule
         Span Enumeration& 94.7 & 52.6 & 72.3 & 51.5 & 28.1 & 17.6& 10.7 \\
         \quad - w/o Typing & \posr{+0.1} & \negr{\textminus0.2} & \negr{\textminus0.1} & \posr{+0.8} & \posr{+1.5} & \posr{+1.3}& \posr{+1.0}\\
      \midrule
         Boundary Prediction & 95.1 & 70.1 & 72.4 & 69.2 & 42.2 & 35.9& 37.6\\
         \quad - w/o Typing & \negr{\textminus0.0}& \negr{\textminus16.9} & \posr{+2.1} & \negr{\textminus9.2} & \negr{\textminus12.6} & \negr{\textminus16.1} & \negr{\textminus27.3}\\

      \bottomrule
    \end{tabular}
    }
    \captionsetup{width=1\linewidth}
    \caption{Experiment results of typing on IE tasks. Positive impact on model performance is shown in \posr{green} while negative in \negr{red}.}
    \label{tab:oneie_typing_exps}
\end{minipage}
\end{table*}

\begin{table*}[t]
\resizebox{\textwidth}{!}{%
    \centering
    \begin{tabular}{c|c|ccccccc|c|cc}
      \toprule
         \bf Encoder & \bf Method & \multicolumn{2}{c}{\bf Entities} & \multicolumn{4}{c}{\bf EE} & \bf RE & \bf CR & \multicolumn{2}{c}{\bf QA} \\
         \cmidrule(lr){3-4}\cmidrule(lr){5-8}\cmidrule(lr){9-9}\cmidrule(lr){10-10}\cmidrule(lr){11-12}
        & & Ent-I & Ent-C & Trig-I & Trig-C & Arg-I & Arg-C & F1 & Avg. F1 & EM & F1 \\
      \midrule
\multirow{3}{*}{RoBERTa\textsubscript{base}}& Tagging & \negr{\textminus0.0}  &  \negr{\textminus1.0} & \negr{\textminus1.3} &  \negr{\textminus2.3} & \negr{\textminus2.9} & \negr{\textminus2.2} & \negr{\textminus1.2} & \posr{+0.9} & \posr{+0.3} & \posr{+0.5}\\
& Span Enumeration & \posr{+0.3} & \posr{+0.1} & \posr{+0.1} & \negr{\textminus0.6} & \posr{+0.8} & \negr{\textminus0.8} & \negr{\textminus1.3} & \negr{\textminus0.1} & \negr{\textminus0.1} & \posr{+0.1} \\
& Boundary Prediction & \posr{+0.5} & \posr{+0.6} & \negr{\textminus0.3} & \posr{+0.6} & \negr{\textminus0.5} & \negr{\textminus0.2} & \negr{\textminus0.8} & \posr{+0.6} &  \posr{+0.5} &  \posr{+0.4}\\
\midrule
\multirow{3}{*}{$\rm T5_{base}^{enc}$}& Tagging & \negr{\textminus0.4} & \posr{+1.0} & \negr{\textminus7.0} & \posr{+9.1} & \posr{+5.2} & \posr{+6.5} &\posr{+4.0} & \posr{+0.1} & \posr{+4.0} & \posr{+3.4} \\
& Span Enumeration & \posr{+1.4} & \posr{+1.7} & \posr{+1.3} & \posr{+5.1} & \negr{\textminus1.6} & \negr{\textminus0.8} & \negr{\textminus0.3}  &  \posr{+0.9} & \posr{+1.3} & \posr{+1.0}\\
& Boundary Prediction & \negr{\textminus0.5} & \negr{\textminus0.8} & \negr{\textminus3.0} &  \negr{\textminus4.0} &  \negr{\textminus1.6} & \negr{\textminus1.9} & \negr{\textminus1.4} & \posr{+1.1} & \posr{+3.6} & \posr{+3.8}\\
\bottomrule
    \end{tabular}
    }
    \caption{Experiment results adding BiLSTM contextual layer to our baseline models. The table shows the performance gaps compared to counterparts in \autoref{tab:major_exps} that do not have an additional contextualization.  Positive impact on model performance is shown in \posr{green} while negative in \negr{red}.}
    \label{tab:lstm_exps}
\end{table*}

\section{Discussions}

\subsection{Which Span Finder to Use?}

From the results presented in \autoref{tab:major_exps}, we find that  boundary prediction is potentially preferable to  span enumeration. 
Although span enumeration outperforms boundary prediction by a small margin in coreference resolution, boundary prediction outperforms span enumeration in other downstream tasks. 
While tagging and span enumeration suffer a considerable performance drop in all downstream tasks when using $\rm T5_{base}^{enc}$ as encoder,  boundary prediction only suffers a slight performance drop. 

From the breakdown of the mention scores of each model on the OntoNotes dataset (coreference resolution) in \autoref{tab:mention_breakdown}, we can see that although span enumeration focuses significantly on recall,\footnote{~In the evaluation of coreference resolution, singleton mention clusters (i.e., clusters that have only 1 mention) are ignored in computing the evaluation scores. This practice weeds out lots of spans that should not be selected as mentions. This is the reason that a coreference model can achieve state-of-the-art results with low mention detection precision.} boundary prediction can reach a comparable level of recall, whereas in a downstream task like QA where precision is needed,  span enumeration cannot reach a comparable level of performance to boundary prediction.

The choice between tagging and boundary prediction depends on various factors, including but not limited to the language model, downstream task, training strategy, etc. Overall, tagging excels at precision; in contrast, boundary prediction and enumeration have better recall.

\subsection{Does Typing Help Span Finders?}

As can be seen from the results in \autoref{tab:oneie_typing_exps}, tagging with untyped labels outperforms tagging with typed labels in all tasks except trigger classification; the margin is even more significant for the tasks of event argument extraction and relation extraction. We hypothesize that under joint training, the types in the labels might hinder the model from learning other objectives. Therefore, if it is not necessary to have typed tags, it is recommended to use plain \texttt{BIO} labels for tagging.

As for boundary prediction, we found that performing classification with types is crucial in the joint training with downstream tasks. This is possibly due to the nature of the two-step process of the method, where the first step can be seen as a coarse classifier to select potential mention candidates while the second step double-checks such candidates (or classifies candidates with more fine-grained types). 
However in span enumeration we observed that it is not much impacted by the inclusion of types. We hypothesize this results from label imbalance. Under span enumeration, there are a considerable number of spans that should not be labelled as valid mention spans. When downstream tasks further require classifying spans to more fine-grained types, the label distribution would be seriously imbalanced and dominated by the $\varepsilon$ (null type) label, making the learning ineffective. 

\subsection{Additional Contextualization?}

We further examine a commonly seen practice of having an additional contextualization with RNN layers atop Transformer encoders. Following \cite{lee-etal-2017-end} as implemented in AllenNLP \citep{gardner-etal-2018-allennlp}, we use a 1-layer BiLSTM with hidden dimension of 200 for each direction. 

We stack the BiLSTM contextual layer atop the Transformer encoders and report the experimental results in \autoref{tab:lstm_exps}. We observe that, for IE tasks, adding additional contextualization does not affect model performance; for extractive QA, it improves model performance. 
Also, when using encoder-decoder architecture models (e.g., T5), the additional contextualization would lead to higher variance in downstream tasks compared to using encoder-only models (e.g., BERT).
We hypothesize that the difference might come from the underlying architecture in T5, in which the encoder learns to specialize its representation to support the decoder. Therefore, when the encoder is being used alone, an additional contextualization might serve a similar purpose as a decoder and have to learn to utilize the representation to some extent. Even with such exceptions, from the design choice perspective, the merit of this trick is limited as it is not helpful in most cases while introducing training variance and additional parameters to the model.





\section{Conclusions}
We identified and investigated three common span finding methods in the NLP community: tagging, span enumeration, and boundary prediction. Through extensive experiments, we found that there is not a single recipe that is best for all scenarios. The design choices on downstream tasks rely on specific task properties, specifically the trade-off between precision and recall. We suggest that precision-focused tasks consider tagging or boundary prediction, and recall-focused (such as coreference resolution) tasks consider span enumeration or boundary prediction. 

We further examined two commonly used tricks to improve span finding performance, i.e., adding span type information during the training and adding additional contextualization with an RNN on top. We observed that boundary prediction on IE tasks heavily relies on type information, and adding additional contextualization mostly does not help span finding. 

Architectures will continue to evolve and models will continue to grow in size, which may lead to different conclusions on the relative benefits of approaches.  Still, the fundamental task of isolating informative spans of text will remain. We hope this study helps inform system designers today with existing models and still in the future as a starting point for further inquiry.

\section{Limitations}
As an empirical study, we provide observations under different combination of design choices for building an end-to-end trained IE systems with a span finding component. We hope such observations could provide insights for future work, but we have to admit that we are also bounded by the limitations of empirical studies, and that a theoretical analysis is out of scope of this paper.
As a result, we only make our claims based on the experiment results with the baseline models that we evaluated, and hope that it could generalize well to other architectures.

\section*{Acknowledgments}

We thank Elias Stengel-Eskin, William Gantt, and Reno Kritz for helpful comments and feedback. This work was supported in part by DARPA AIDA (FA8750-18-2-0015) and IARPA BETTER (2019-19051600005). The views and conclusions contained in this work are those of the authors and should not be interpreted as necessarily representing the official policies, either expressed or implied, or endorsements of DARPA or the U.S. Government. The U.S. Government is authorized to reproduce and distribute reprints for governmental purposes notwithstanding any copyright annotation therein.

\bibliography{anthology,custom}
\bibliographystyle{acl_natbib}






\end{document}